\documentclass[twoside,11pt]{article}
\usepackage{mhStyle}
\usepackage{amssymb,amsmath,times}
\usepackage[ruled,vlined]{algorithm2e}
\usepackage{booktabs,balance}
\usepackage{xspace,url,subcaption,multirow}
\usepackage[backref = false,colorlinks=true,linkcolor=blue,citecolor=green,urlcolor=blue]{hyperref}

\usepackage{url,xspace,caption,subcaption}
\usepackage{multirow}
\usepackage{array}

\graphicspath{{./}{./figures/}}

\makeatletter
\DeclareRobustCommand\onedot{\futurelet\@let@token\@onedot}
\def\@onedot{\ifx\@let@token.\else.\null\fi\xspace}

\def\eg{\emph{e.g}\onedot} 
\def\ie{\emph{i.e}\onedot}

\def\etal{\emph{et al}\onedot}
\makeatother

\def\Vec#1{{\boldsymbol{#1}}}
\def\Mat#1{{\boldsymbol{#1}}}

\def\ST#1#2{\mathrm{St}({#2},{#1})}
\def\NCST#1#2{\mathrm{St}^\ast({#2},{#1})}

\newcommand{\DIAG}{\mbox{Diag\@\xspace}}
\newcommand{\subject}{\mathrm{s.t.}}

\newcolumntype{K}[1]{>{\centering\arraybackslash}m{#1}}

\begin{document}

\title{Generalized BackPropagation\\ \'{E}tude De Cas: Orthogonality}

\author
  {
  \name Mehrtash Harandi \email {mehrtash.harandi@anu.edu.au} \\
  \addr Data61, CSIRO, Canberra, Australia \\
  \addr Australian National University, Canberra, Australia \\ 
  \name Basura Fernando \email {basura.fernando@anu.edu.au} \\
  \addr Australian National University, Canberra, Australia \\
  }

\maketitle

\begin{abstract}

This paper introduces an extension of the backpropagation algorithm that enables us to 
have layers with constrained weights in a deep network. 
In particular, we make use of the Riemannian geometry and optimization techniques on matrix manifolds 
to step outside of normal practice in training deep networks, equipping the network with 
structures such as orthogonality or positive definiteness.
Based on our development, we make another contribution by introducing the Stiefel layer,
a layer with orthogonal weights. Among various applications, Stiefel layers
can be used to design orthogonal filter banks, perform dimensionality reduction and feature extraction.
We demonstrate the benefits of having orthogonality in deep networks through a broad set of experiments,
ranging from unsupervised feature learning to fine-grained image classification.
\end{abstract}

\section{Introduction}
\label{sec:intro}

In this paper, we discuss how deep networks with structured layers can be trained
and study properties attained by one especial construction, namely  orthogonality.
In doing so, we introduce the generalized BackPropagation (gBP) algorithm
that benefits from the Riemannian optimization techniques to maintain structural properties envisaged in layers
of a deep network.

A deep network, as a machinery, 
is an elaborated and hierarchical composition of simple functions.
Each function is realized through a layer with a set of trainable parameters
or weights. Enforcing structure\footnote{
In this paper, by structured-layer, we mean a layer when some form of structure is envisaged on the weights.
} (\eg, orthogonality) on the weights of a network may lead to better generalization abilities as it constraints the parameter space. 
This naturally raises two questions, \textbf{1.} What type of constraints are useful in deep networks? and 
\textbf{2.} How the parameters of a constrained-layer can be learned? We touch upon the first question based on 
the desired properties of orthogonality and provide a simple, yet mathematically rigorous answer to the second question.

BackPropagation (BP)~\cite{Bryson_1969,Rumelhart_1986,Lecun_BP_2012}, a gradient-based learning method,  is 
the most popular choice for training deep networks, nowadays. However and as will be shown shortly, 
the updating scheme of BP is incapable of preserving 
constraints envisaged for the weights of a layer. This limitation of BP motivates us to develop the gBP algorithm,
an extension of BP that can accommodate constraints on the weights of deep architectures.

By making use of the gBP, we introduce our next player in the paper, the Stiefel\footnote{%
Eduard Stiefel (1909 -– 1978) was a Swiss mathematician who together with Cornelius Lanczos and Magnus Hestenes invented the  
conjugate gradient method in 1952. The Stiefel manifold, the geometry encompassing the set of orthogonal matrices,
is named in honor of him. We also name layers with orthogonal weights after him.}
layer, a layer with orthogonal weights. Among various applications, Stiefel layers can be used to design orthogonal filter banks, 
perform dimensionality reduction and feature extraction.
The concept of orthogonality is essential in learning theory. 
Examples include but not limited to 
dimensionality reduction~\cite{Turk_1991,Hyvarinen_1999,Cai_TIP_2006,Cunningham_JMLR_2015}, 
clustering~\cite{Niu_ICML_2010,Kumar_2011},
feature selection~\cite{Yu_2004} and  
dictionary learning~\cite{Lesage_05,Vidal_2005}.
However, and to a great extent surprising, orthogonality is a topic barely touched in the 
literature of deep learning.
Our work bridges this gap and opens the door to unexplored venues.

\subsection*{Contributions.}
\noindent
Our contributions in this work can be summarized as:

\textbf{1.} We introduce the generalized backpropagation algorithm. This enables us to design Stiefel layers, layers with orthogonal weights. 
\textbf{2.} We make use of Stiefel layers to learn features from face images in an unsupervised fashion.
\textbf{3.} We use the Stiefel layers for the purpose of image classification. In particular, we compare and
analyze the benefits achieved by having orthogonality in the LeNet~\cite{Lecun_1995} over CIFAR10~\cite{CIFAR_DB},
CIFAR100~\cite{CIFAR_DB}, and STL~\cite{STL_DB} datasets.
\textbf{4.} We incorporate Stiefel layers with AlexNet~\cite{Krizhevsky_NIPS_2012} and various structures of VGGNet~\cite{Simonyan_2014}  to perform fine-grained classification. In particular, we assess the classification accuracies of deep nets with 
Stiefel layers on birds~\cite{CUB200_DB}, aircrafts~\cite{AIRCRAFT_DB} and cars~\cite{CARS196_DB} datasets.
\textbf{5.} We make use of the Stiefel layers to simplify a deep net through low-rank approximations.

Before delving into technical details, 
we provide a sneak pick at some of the obtained results.
Compared to their generic counterparts, we will show that having orthogonality improves the classification accuracy of,
\textbf{1.}  the LeNet~\cite{Lecun_1995} from 51.4\% to 62.1\% on the STL dataset~\cite{STL_DB},
\textbf{2.}  the AlexNet~\cite{Krizhevsky_NIPS_2012}  from 68.4\% to 70.5\% on the 
CUB200~\cite{CUB200_DB} dataset,
\textbf{3.}  the VGG-VD~\cite{Simonyan_2014} from 86.0\% to 87.9\% on theCars196~\cite{CARS196_DB} dataset.
We will see that on the Cars196~\cite{CARS196_DB} dataset, 
replacing the fc7 layer of the VGG-M~\cite{Simonyan_2014} by orthogonal layers leads to 
boosting the classification accuracy from 77.5\% to 82.0\%, 
while the number of parameters of fc7 is reduced from 16.7M to 745K.  
The Matlab code to perform gBP along the Stiefel layer designed to be incorporated in MatConvNet package~\cite{matconvnet} 
are available at~\url{https://sites.google.com/site/mehrtashharandi/}. 

\section{Generalized BackPropagation}
\label{sec:gbp}

In this section, we present the gBP algorithm, and follow it up by providing the specific form of gBP to incorporate
orthogonality constraints in deep networks. 

\vspace{2ex}
\noindent
\textbf{Notations.}
~Throughout the paper, bold capital letters denote matrices (\eg, $\Mat{X}$) and bold lower-case letters denote column vectors (\eg, $\Vec{x}$).
$\mathbf{I}_n$ is the $n \times n$ identity matrix.
$\mathrm{GL}(n)$ denotes the  general linear group, the group of real invertible $n \times n$ matrices.
We denote the set of full-rank $n \times p$ matrices by $\mathbb{R}_\ast^{n \times p}$.
The orthogonal group is denoted by $\mathcal{O}_n$, \ie, 
$\mathcal{O}_n = \{\Mat{R} \in \mathbb{R}^{n \times n}| \Mat{R}\Mat{R}^T = \Mat{R}^T\Mat{R} = \mathbf{I}_n\}$.
$\mathrm{Diag} \left( \lambda_1,\lambda_2,\cdots,\lambda_n \right)$ is a diagonal matrix formed from
real values $\lambda_1,\lambda_2,\cdots,\lambda_n$ on diagonal elements.

\subsection{BackPropagation}
\label{sec:closer_look_bp}

Consider a learning problem over $\mathbb{X} = \{\Vec{x}_i\}_{i = 1}^N, \Vec{x}_i \in \mathbb{R}^n$
using a multilayer feed-forward network with $K$-layers.
The network is parameterized 
by a set of weights $\mathbb{W} = \{\Mat{W}_i\}_{i = 1}^K$, biases $\mathbb{B} = \{\Vec{b}_i\}_{i = 1}^K$ 
and non-linearities $\{f_i\}_{i = 1}^K$ and realizes a class of functions in the form $f = f_i \circ f_{i-1} \circ \cdots \circ f_1$.
Let $\mathbb{D} = \{\Vec{d}_i\}_{i = 1}^N, \Vec{d}_i \in \mathbb{R}^C$ be the set of desired target values associated 
to the samples in $\mathbb{X}$. Consider $\ell:\mathbb{R}^C \times \mathbb{R}^C \to  \mathbb{R}$ to be a loss function,
measuring the mismatch between the \emph{prediction of the network} $f:\mathbb{R}^n \to \mathbb{R}^C$ and the desired output.
The empirical loss of the network
is defined as
\begin{equation*}
E = \frac{1}{N}\sum_{i=1}^N \ell \big( f(\Vec{x}_i,\mathbb{W},\mathbb{B}),\Vec{d}_i \big).
\end{equation*}

In its simplest form, the BP algorithm updates the weights of a layer according to
\begin{align}
\Mat{W}^{(t)} = \Mat{W}^{(t-1)} - \eta \frac{\partial E}{\partial \Mat{W}}\;,
\label{eqn:bp_update_weight}
\end{align}
with $\eta$ being the learning rate\footnote{ 
We have dropped the layer subscripts to increase the readability. 
}. 
The underlying assumption of BP is that the geometry of the parameter space is Euclidean.
This innocent looking assumption prevents BP from preserving structures envisaged in a network.
As an example and inline with the focus of this paper, suppose an orthogonal layer (\ie, $\Mat{W}^T\Mat{W} = \mathbf{I}_p$) is required in a network. Clearly, there is no guarantee that updates according to Eq.~\eqref{eqn:bp_update_weight} keep  
the orthogonality structure.

In contrast to BP, the gBP algorithm  can accommodate certain constraints on the weights of a network. 
Orthogonality is arguably the first and most important constraint to study 
and will be treated in depth in \textsection~\ref{sec:stiefel_layer}. Aside from orthogonality, other constraints that gBP can bring into 
the architecture of a network include but not limited to 
\begin{enumerate}

\item \textbf{Subspace Constraint.} Another form of the orthogonality constraint is the case when invariance to the basis of a subspace 
	   is required. Formally, we want to have $\Mat{W}^T \Mat{W} = \mathbf{I}_p$ with an extra condition that all the elements of the set 
	   $\mathbb{W} \triangleq \{\Mat{W}\Mat{R}\},~\Mat{R} \in \mathcal{O}_p$ being equivalent for the optimization. 
	   A good example here is a metric learning problem. Let the empirical loss be $E \triangleq \sum \ell(\| \Vec{y}_i - \Vec{y}_j \|^2)$ 
	   with $\Vec{y} = \Mat{W}^T\Vec{x}$ being the output of a fully-connected (fc) layer placed before the loss layer. 
	   Assume an orthogonality is envisaged on $\Mat{W}$, we note that 
	   \begin{align*}
	   g(\| \Vec{y}_i - \Vec{y}_j \|^2) &= g(\| \Mat{W}^T\Vec{x}_i - \Mat{W}^T\Vec{x}_j \|^2) \\
	   &=  g(\| \Mat{R}\Mat{W}^T\Vec{x}_i - \Mat{R}\Mat{W}^T\Vec{x}_j \|^2),
	   \end{align*}
	   As such, orthogonality constraint for this problem is indeed a subspace constraint. 
	   The subspace constraint can be added to the layers of a network by 
	   making use of the Grassmannian geometry~\cite{Edelman_1998,Absil_Book} in gBP.

\item \textbf{Positive-Definiteness.} We deem $\Mat{W} \in \mathbb{R}^{n \times n}$ to be positive definite (p.d.), 
	  \ie, $\Vec{x}^T\Mat{W} \Vec{x} > 0,~\forall \Vec{x} \in \mathbb{R}^n - \{\Vec{0}\}$. This constraint is useful in learning metrics,
	  correlation and kernel matrices among the others. The p.d. constraint can be added to the layers of a network by 
	   exploiting the geometry of the Symmetric Positive Definite (SPD) manifolds~\cite{Pennec_IJCV_2006} in gBP.

\item \textbf{Equal Energy Filters.} We deem $\Mat{W} \in \mathbb{R}^{n \times p}$ to have unit norm columns. 
	  This constraint, when used in convolutional layers, leads to having equal energy filters, a property frequently used in signal processing. 
	  This constraint can be incorporated into convolutional layers by making use of the geometry of the 
	  oblique manifolds~\cite{Absil_ICASSP_2006}  in gBP.

\end{enumerate}

\subsection{From Euclid to Riemann.}
\label{sec:euclid_riemann}

To preserve some form of constraints during BP, one may opt for the method of 
Projected Gradient Descent (PGD)~\cite{Boyd_Book}.  
In PGD, optimization is proceed by projecting the gradient-descent updates onto the set of constraints. For example, in 
the case of orthogonality, after an update by ignoring the orthogonality constraint, the result is projected 
onto the set of orthogonal matrices. 

In PGD, to perform the projection, the set of constraints needs to be closed. In practice, however one can resort to 
open sets. For example, the set of SPD matrices is open though one can project a symmetric matrix onto this set using eigen-decomposition.
In this work, we propose a more principle approach to preserve the constraints by making use of the Riemannian optimization technique. 
We take a short detour and describe the Riemannian Gradient Descent (RGD) method below.

\subsection*{Riemannian Gradient Descent.}

Consider a non-convex and constrained optimization problem in the form of 
\begin{align}
\min_{\Mat{W} \in \mathcal{M}}~E(\Mat{W})
\;.
\label{eqn:opt_riemannian_manifold}
\end{align}
Assume $\mathcal{M}$, the set of constraints, forms a Riemannian manifold. Informally,  
a Riemannian manifold is a smooth surface, equipped with a metric to measure lengths and angles which resembles Euclidean spaces locally. Many constraints of interest in the learning theory
such as orthogonality and positive definiteness fulfill our assumption here, \ie, they form Riemannian manifolds. 
To optimize~\eqref{eqn:opt_riemannian_manifold}, RGD improves 
its solution iteratively according to 
\begin{equation}
\Mat{W}^{(t+1)} = \Upsilon\Big( -\eta ~\mathrm{grad}~E(\Mat{W}^{(t)})\Big)\;.
\label{eqn:rgd_updates}
\end{equation}

\begin{figure*}[t!]
\centering
\begin{subfigure}{.45\textwidth}
	\centering
	\includegraphics[width = 0.8 \linewidth,keepaspectratio]{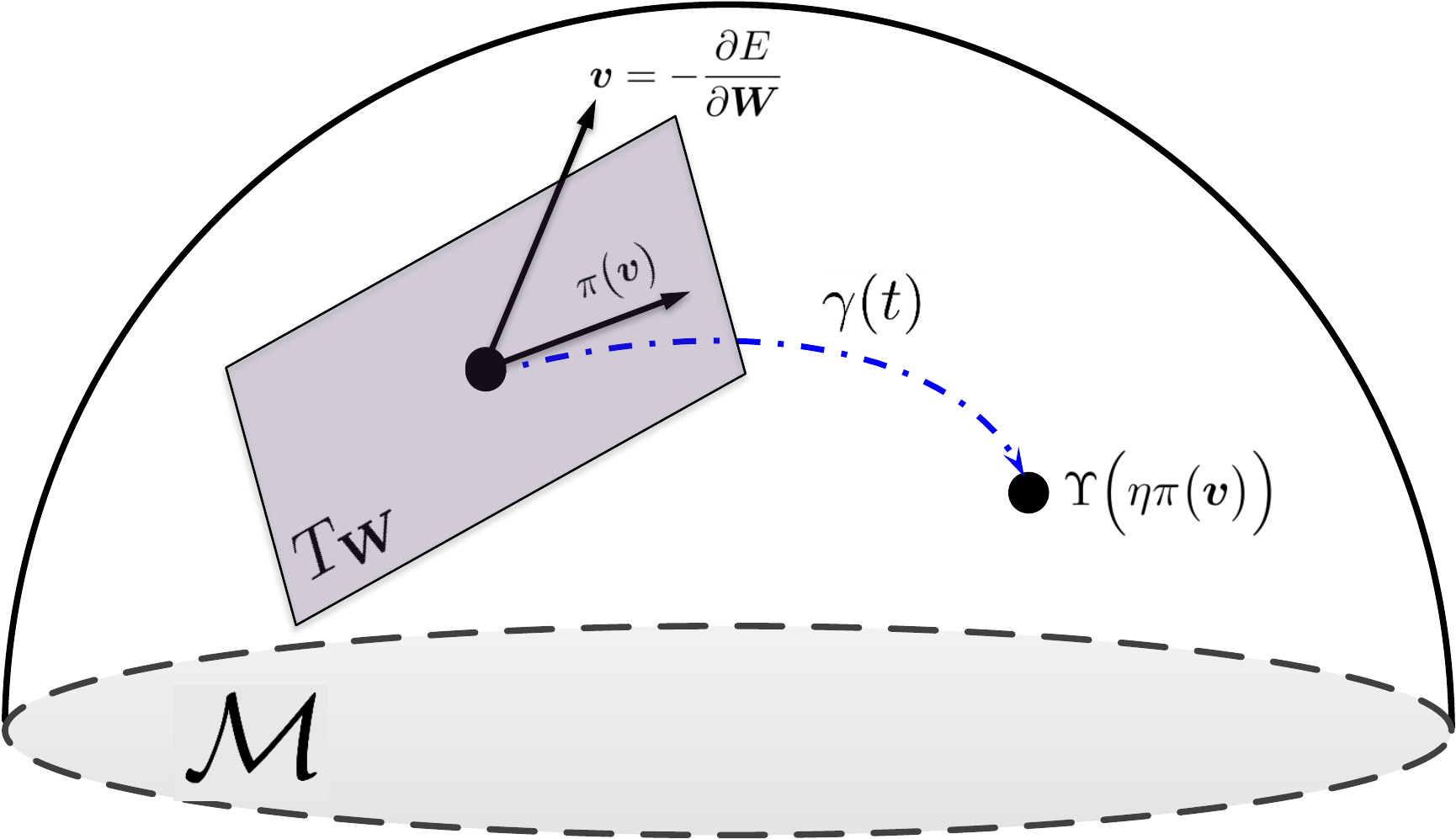}
	\caption{}
	\label{fig:rgd}
\end{subfigure}%
\begin{subfigure}{.45\textwidth}
	\centering
	\includegraphics[width = 0.8 \linewidth,keepaspectratio]{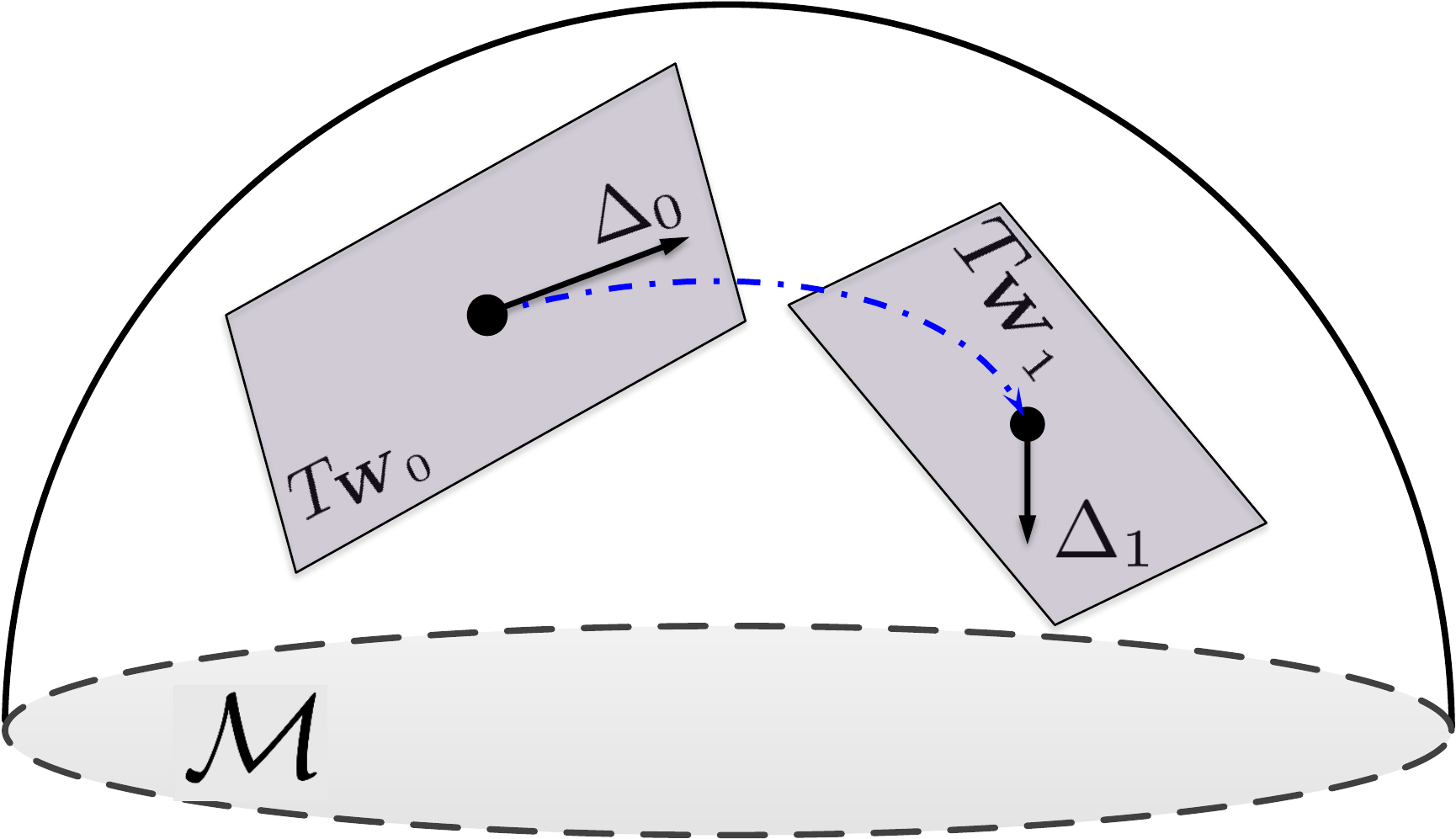}
	\caption{}
	\label{fig:r_connections}
\end{subfigure}%
\caption{\footnotesize \textbf{a)} \textbf{Constrained optimization using Riemannian geometry.} Given the descent direction $\Vec{v}$, 
moving along $\Vec{v}$ takes the gradient descent update off the manifold and results in violating the constrained. On the contrary, in 
optimization on Riemannian manifolds, the descent direction $\Vec{v}$ identifies the Riemannian gradient, an element of the Tangent space 
$T_\Mat{W}$ shown by $\pi(\Vec{v})$.  The Riemannian gradient defines a curve on the manifold whose points can be reached locally by the retraction operator $\Upsilon(\cdot)$.
\textbf{b)} \textbf{The concept of Riemannian connections.} The tangent vector $\Delta_0 \in T_{\Mat{W}_0}$
cannot be directly used in another tangent space $T_{\Mat{W}_1}$. $\Delta_1$ is the transport of $\Delta_0$ using Riemannian connections.
This is important to extend the notion of momentum in gBP.
}
\label{fig:riemannian_concepts}
\end{figure*}

In Eq.~\eqref{eqn:rgd_updates}, $\eta > 0$ is the RGD step size, $\mathrm{grad}~E(\cdot)$ is the Riemannian gradient of the objective function, and $\Upsilon ( \cdot): T_\Mat{W} \to \mathcal{M}$
is called a \emph{retraction}, with $T_\Mat{W}$ being the tangent space of $\mathcal{M}$ at $\Mat{W}$. 
With the help of Fig.~\ref{fig:riemannian_concepts}, we introduce these concepts at an 
intuitive level. See~\cite{Absil_Book} for a rigorous treatment.

Without considering the constraint set $\mathcal{M}$, ${-\partial E}/{\partial \Mat{W}}$
shown by $\Vec{v} \in \mathbb{R}^n$ in Fig.~\ref{fig:rgd} determines the maximum descent direction in Euclidean geometry. Alas, moving in the  
direction of $\Vec{v}$ makes the new solution off the manifold, resulting in violating the constraints. On the contrary, the Riemannian gradient identifies a curve $\gamma(t)$ on the manifold that ensures to reduce the objective function (at least locally). 
For a large group of Riemannian manifolds, including the ones that are of interest here, 
the explicit form of obtaining Riemannian gradient from ${\partial E}/{\partial \Mat{W}}$ is known. 
This is shown by the operator $\pi:\mathbb{R}^n \to T_\Mat{W}$ in Fig.~\ref{fig:rgd}\footnote{
Generally speaking, $\pi$ depends on the point of projection. Here, as our goal is to provide a high-level understanding of RGD, 
we omit such dependencies in our notations.}.

To obtain points on $\gamma(t)$ from the Riemannian gradient, a Riemannian operator, namely, exponential map is required. 
Computing exponential maps, if possible, is computationally expensive in most cases. Instead, in Riemannian optimization, 
exponential maps are approximated using retractions $\Upsilon:T_\Mat{W} \to \mathcal{M}$. 
A retraction serves to jointly move in the descent direction and guarantees that the new solution is on the 
manifold $\mathcal{M}$. Obviously, the form of $\pi$ and $\Upsilon$ is manifold-specific.


We note that Riemannian optimization techniques come with convergence guarantees, essentially matching the
Euclidean counterparts they generalize. For example, the Riemannian trust-region
method converges globally towards critical points~\cite{Absil_Book}.
In its simplest form, gBP updates the weights of a layer with constraints according to 
\begin{align}
\Mat{W}^{(t)} &= \Upsilon \Big( - \eta \mathrm{grad} E \Big)
&= \Upsilon \Big( - \eta \pi\Big(\frac{\partial E}{\partial \Mat{W}}\Big) \Big)\;.
\label{eqn:gbp_update_weight}
\end{align}

\begin{remark}
For the Euclidean geometry, the Riemannian form of $\pi(\cdot)$ and  $\Upsilon(\cdot)$ at a given point $\Mat{W}$
are given by $\pi(\Mat{A}) = \Mat{A}$ and $\Upsilon(\Mat{A}) = \Mat{W} + \Mat{A}$, respectively. Plugging this back to the gBP,
we will see that BP updating rule is recovered, hence BP is a special case of gBP when the geometry is Euclidean.
\end{remark}

\subsection*{Momentum}
\label{subsec:momentum}
The momentum method~\cite{Polyak_1964}, a technique to accelerate the convergence of the gradient descent,
is widely used in BP~\cite{Lecun_BP_2012,Sutskever_ICML_2013}. The classical form of momentum reads
\begin{align}
\label{eqn:euclid_momentum}
\Theta^{(t)}    &= \mu \Theta^{(t-1)} - \eta \frac{\partial E}{\partial \Mat{W}^{(t-1)}}\;.
\\
\Mat{W}^{(t)} &= \Mat{W}^{(t-1)} + \Theta^{(t)}\;,
\label{eqn:euclid_momentum_update}
\end{align}
with $\mu \in [0,1]$ being the momentum coefficient. In Eq.~\eqref{eqn:euclid_momentum},
$\Theta^{(t)}$ can be seen as a smoother form of gradient, helping the gradient descent to move away 
from low-curvature surfaces faster. 
To extend the concept of momentum to Riemannian setting, the notion of Riemannian connections is required. 
To  put this into perspective, consider the second run of Eq.~\eqref{eqn:euclid_momentum} given by
\begin{align}
\Theta^{(2)}    
				 &= -\mu \eta \frac{\partial E^{(1)}}{\partial \Mat{W}^{(0)}} - \eta \frac{\partial E^{(2)}}{\partial \Mat{W}^{(1)}}\;.
\label{eqn:momentum_2nd_update}
\end{align}

That is, $\Theta^{(2)}$ is obtained by adding two gradients computed at different locations. 
In contrast to the Euclidean spaces, tangent vectors residing on different tangent spaces cannot be 
readily added together (see Fig.~\ref{fig:r_connections} for an illustration). Theoretically, one needs to use  
\emph{Riemannian connections}~\cite{Absil_Book} to transport 
a tangent vector to its target tangent space before performing any addition. 
Generally speaking, Riemannian connections are computationally expensive. Therefore, it is a well-practiced\footnote{The 
Manopt package~\cite{manopt} extensively uses this idea to perform optimization on Riemannian manifolds.} idea to 
make use of the projection $\pi(\cdot)$ to simplify the computations when it comes to Riemannian optimization methods. 
Based on the discussion above, we propose the following updating scheme for gBP with momentum

\begin{align}
\Theta^{(t)}    &= \mu \Theta^{(t-1)} - \eta \frac{\partial E}{\partial \Mat{W}^{(t-1)}}\;.
\notag\\
\Mat{W}^{(t)} &= \Upsilon \Big( \pi\big(\Theta^{(t)}\big) \Big)\;.
\label{eqn:gbp_update_weight_momentum}
\end{align}

An in-depth study over other forms of smoothing and their associated properties for gBP demands a dedicated work
and goes beyond the scope of our paper.

We discussed that the PGD method can also be used to add constraints on the weights of a deep network. Nevertheless,
we propose a Riemannian extension of BP as the method of choice. A curious mind may ask why not PGD? 
We note that the gBP has a simple, yet very general form for preserving structures. Furthermore, gBP can be incorporated 
into deep learning packages with the minimum effort (the SGD module should accept two extra functions for each layer to 
perform the projection $\pi$ and the retraction $\Upsilon$). 

To contrast PGD from gBP, we performed a simple test by considering the problem of identifying the eigenvectors 
of a covariance matrix.
In particular, consider a network realizing $f(\Vec{x}) = \Mat{W}^T\Vec{x}$ with  $\Mat{W} \in \mathbb{R}^{n \times p}$,
$\Vec{x} \in \mathbb{R}^n$ and the orthogonality constraint on $\Mat{W}$, \ie, $\Mat{W}^T\Mat{W} = \mathbf{I}_p$. 
Define the network loss to be 

\begin{equation*}
E = \frac{1}{N}\sum_{i=1}^N \Big\| \Vec{x}_i -  \Mat{W}\Mat{W}^T \Vec{x}_i \Big \|^2.
\end{equation*}

\begin{figure*}[t!]
	\centering
	\includegraphics[width = 0.6 \linewidth,keepaspectratio]{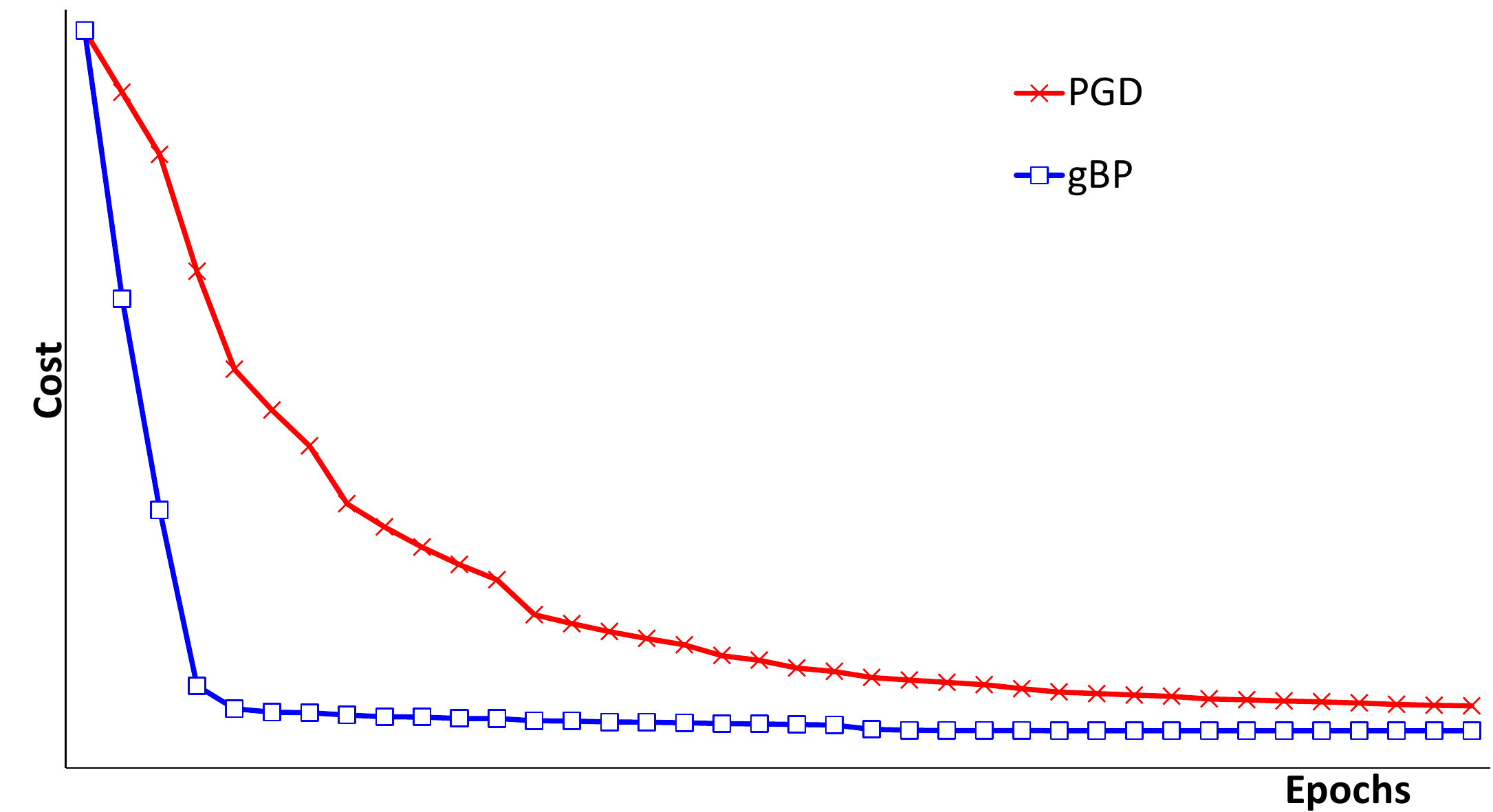}
	\caption{\footnotesize 
	\textbf{Convergence behavior of gBP and PGD for approximating eigenvectors of a matrix.} gBP not only converges faster for this 
	problem but also it achieves a lower loss.
}
\label{fig:pdg_bgp}
\end{figure*}

First we note that the form $\Mat{W}\Mat{W}^T \Vec{x}$ is the reconstruction of $\Vec{x}$ using the subspace
spanned by $\Mat{W}$. If the data is centered (\ie, $\sum_i \Vec{x}_i = \Vec{0}$), then with the orthogonality constrain, 
the network should identify the top $p$ eigenvectors of $\sum_i \Vec{x}_i\Vec{x}_i^T$ after training. 
In Fig.~\ref{fig:pdg_bgp}, we plot the evolution of the loss function during training for the PGD and gBP methods.
We note that the gBP algorithm not only converges  faster but also it achieves a smaller loss. While drawing a solid conclusion 
from this experiment is not our intention, the performance of gBP in comparison to PGD is pleasing.

\begin{remark}
In gBP, we did not discuss how  biases will be updated, and 
how ${\partial E}/{\partial \Mat{W}}$ can be obtained for a layer.
This is because the BP practice applies verbatim to gBP. That is, the gradient ${\partial E}/{\partial \Mat{W}}$
is obtained through backward recurrence and no changes to the BP algorithm are required. Similarly for the biases, since 
we do not envisage any constraints on them, the BP updates follow.
\end{remark}

Before concluding this part, 
We emphasize that though we are chiefly interested in  orthogonality in this work, the gBP algorithm 
can be used to incorporate various structures (\eg, positive definiteness) into a network. 
Furthermore, we note that our development above can be applied verbatim to train recurrent networks by modifying the 
BackPropagation Through Time (BPTT)~\cite{Werbos_1990} algorithm accordingly.


\section{Stiefel Layers}
\label{sec:stiefel_layer}

In this section, we adapt the gBP algorithm to enforce orthogonality on the weights of a deep network. 
This is achieved by making use of the geometry of the Stiefel manifold~\cite{Absil_Book}. To do complete justice, let us formally define the Stiefel manifold.

\begin{definition}[The Stiefel Manifold]
The set of ($n \times p$)-dimensional matrices, $p \leq n$, with orthonormal columns endowed with the Frobenius inner product\footnote{%
Note that the literature is divided between this choice and another form of Riemannian metric. 
See~\cite{Edelman_1998} for details.} forms a compact Riemannian manifold
called the  Stiefel manifold $\ST{n}{p}$~\cite{Absil_Book}.

\begin{equation}
\ST{n}{p} \triangleq \{\Mat{W} \in \mathbb{R}^{n \times p}: \Mat{W}^T\Mat{W} = \mathbf{I}_p\}\;.
\label{eqn:stiefel_manifold}
\end{equation}
\end{definition}

As discussed in \textsection~\ref{sec:gbp}, the gBP algorithm requires the form of Riemannian gradient and a retraction on $\ST{p}{n}$. 
For $f:\ST{n}{p} \to \mathbb{R}$, the Riemannian gradient is obtained as~\cite{Absil_Book}

\begin{equation}
\label{eqn:stiefel_grad}
\mathrm{grad}_\Mat{W}\big( E \big) = \frac{\partial E}{\partial \Mat{W}}  
- \Mat{W}\mathrm{sym}\Big(\Mat{W}^T \frac{\partial E}{\partial \Mat{W}} \Big)\;.
\end{equation}

In Eq.~\eqref{eqn:stiefel_grad}, $\mathrm{sym}(\Mat{A}) = \frac{1}{2}(\Mat{A} + \Mat{A}^T)$.
Various forms of retraction are defined on $\ST{n}{p}$~\cite{Absil_Book}. Among them, we recommend the following retraction
\begin{equation}
\label{eqn:stiefel_retraction}
\Upsilon_\Mat{W}\big( \xi \big) = \mathrm{qf}(\Mat{W} + \xi)\;.
\end{equation}
Here, $\mathrm{qf}(\Mat{A})$ is the adjusted  $\Mat{Q}$ factor of the QR decomposition~\cite{Golub_Book}.
More specifically, let $\mathbb{R}_\ast^{n \times p} \ni \Mat{A} = \Mat{Q}\Mat{R}$ with $\Mat{Q} \in \ST{n}{p}$ 
and $\Mat{R}$ being an upper triangular $p \times p$ matrix with \emph{strictly positive} 
diagonal elements. Then $\Mat{Q} = \mathrm{qf}(\Mat{A})$. In practice, to obtain $\mathrm{qf}(\cdot)$, one performs QR 
decomposition followed by swapping the sign of elements of all columns whose corresponding diagonal elements in $\Mat{R}$ are negative.
Putting Eq.~\eqref{eqn:stiefel_grad} and Eq.~\eqref{eqn:stiefel_retraction} together, 
the updating rule for the Stiefel layers reads as

\begin{align*} \small
\Mat{W}^{(t)} &\hspace*{-1ex}= \mathrm{qf}\bigg(
\eta \Mat{W}^{(t-1)}\mathrm{sym}\Big({\Mat{W}^{(t-1)}}^T \hspace*{-0.5ex}\frac{\partial E}{\partial \Mat{W}} \Big) 
\hspace*{-0.1ex}+ \hspace*{-0.1ex}
\Mat{W}^{(t-1)}\hspace*{-1ex} -\hspace*{-0.5ex} \eta \frac{\partial E}{\partial \Mat{W}} \bigg).
\end{align*}

\begin{remark}[Cayley Transform]
Preserving orthogonality can also be attained using the Cayley transform~\cite{Wen_2013}. 
The Cayley transform is indeed a valid form of retraction on the Stiefel manifold. 
Therefore, updating the weights of a layer using the Cayley transform can be understood as a special form of the gBP algorithm 
customized for the Stiefel geometry. The retraction provided in Eq.~\eqref{eqn:stiefel_retraction} is however computationally 
cheaper and hence preferable. 
\end{remark}

\begin{remark}[Dimensionality]
The dimensionality of $\ST{n}{p}$ is $np - \frac{1}{2}p(p+1)$~\cite{Absil_Book}. 
The reduction in the number of parameters from $np$ to $np - \frac{1}{2}p(p - 1)$ may become substantial 
in computer vision problems. For example, AlexNet~\cite{Krizhevsky_NIPS_2012} and variants of VGG-Net~\cite{Simonyan_2014} 
make use of a fully connected layer of size $4096 \times 1000$ before the softmax layer. Replacing such a layer with an Stiefel
one reduces the number of parameters from 4M to less than 3.6M.
\end{remark}

\subsection*{Computational Complexity.}
The complexity of each iteration of SGD for an Stiefel layer depends on the computational cost of the following major steps:
\begin{itemize}
\item \textbf{Riemannian gradient.} Projecting ${\partial E}/{\partial \Mat{W}}$ to the tangent space of $\ST{n}{p}$ 
as in Eq.~\eqref{eqn:stiefel_grad} involves multiplications between matrices of size \textbf{1-} $n \times p$ and $p \times p$ and \textbf{2-} $p \times n$ and $n \times p$. This adds up to $2np^2$ flops.

\item \textbf{Retraction.} The retraction involves computing and adjusting the QR decomposition~\cite{Golub_Book} of an $n \times p$ matrix. 
The complexity of the QR decomposition using the Householder algorithm is $2p^2(n - p/3)$~\cite{Golub_Book}. 
Adjustments change the sign of elements
of a column if the corresponding diagonal element of $\Mat{R}$ is negative which does not incur much. Hence, the total complexity of  
the retraction is $O\big(2p^2(n - p/3)\big)$. 

\end{itemize}

All the above steps are linear in $n$, making the extra flops compared to the conventional updates in Eq.\eqref{eqn:bp_update_weight}
affordable. Note that all the above operations can be done in a GPU.

\begin{figure}[tb!]
\centering
\includegraphics[width = 0.75 \textwidth,keepaspectratio]{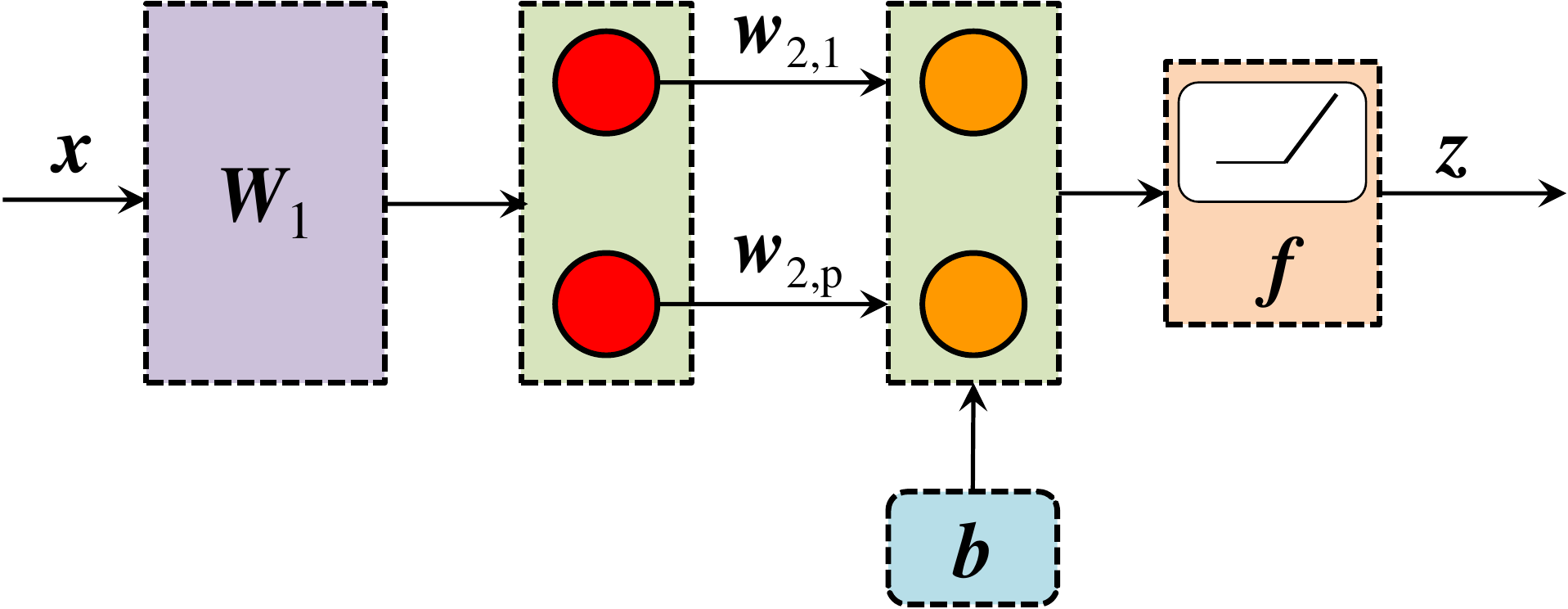}
\caption{\footnotesize \textbf{Non-Compact Stiefel Layer.}  The layer realizes the function 
$\Vec{z} = f\Big(\Mat{W}_2\Mat{W}_1^T\Vec{x} + \Vec{b}\Big)$ with 
$\Mat{W}_1^T\Mat{W}_1 = \mathbf{I}_p$ and $\Mat{W}_2 = \DIAG(w_{2,1},\cdots,w_{2,p})$.}
\label{fig:nc_stiefel}
\end{figure}

\subsection*{Non-Compact Stiefel Layer}

To have a more flexible form of orthogonality in deep networks, we define the non-compact Stiefel layer as a layer whose weights 
satisfy the constraint 
\begin{align}
\NCST{n}{p} &\triangleq \Big\{\Mat{W} \in \mathbb{R}^{n \times p}:\Mat{W}^T\Mat{W} = \mathrm{Diag} \left( \lambda_1,\cdots,\lambda_p \right)
 \Big\}\notag\\
&~\subject~~~ \lambda_i \neq 0,~~ i \in \{1,2,\cdots,p\}\;.
\label{eqn:nc_stiefel_manifold}
\end{align}

From a topological point of view, we can model $\NCST{n}{p}$ as the product topology\footnote{%
Our definition here differs  from the definition in~\cite{Absil_Book} where  the non-compact Stiefel manifold is defined as the set of 
$n \times p$  matrices whose columns are linearly independent. The non-compact Stiefel manifold according to~\cite{Absil_Book} can be understood  as the product space of the Stiefel manifold and the space of full rank $p \times p$ matrices. }  
of $\ST{n}{p}$ and $\mathbb{R}^p- \{\Vec{0}\}$. This can be implemented in a deep network by combining an Stiefel layer with an 
fully connected (fc) layer with one-on-one connections as shown in Fig.~\ref{fig:nc_stiefel}.

\subsection{Low-Rank Factorization of Fully-Connected Layers}
\label{sec:lr_layer_simplification}

Simplifying and pruning deep networks, especially the fc layers, 
is an active line of research~\cite{Han_NIPS_2015,Yang_ICCV_2015_DFC} as such layers are memory intensive 
and computationally demanding. Below we show how Stiefel layers can be used to simplify fc layers in a deep network. 
Let $\mathbb{R}^{n_1 \times n_2} \ni \Mat{W} = \Mat{U} \Mat{D} \Mat{V}^T$ be the SVD factorization of the parameters of an fc layer. A low-rank approximation to $\Mat{W}$ is obtained by $\Mat{W} \approx \Mat{U}_p \Mat{D}_p \Mat{V}_p^T$ where 
$\Mat{U}_p \in \ST{n_1}{p}$ and $\Mat{V}_p \in \ST{n_2}{p}$  are the first $p$ columns 
of $\Mat{U}$ and $\Mat{V}$, respectively. Similarly, $\Mat{D}_p = \DIAG(d_1,d_2,\cdots,d_p)$ stores the first $p$ singular values of $\Mat{W}$.

\begin{figure}[tb!]
\centering
\includegraphics[width = 0.75 \textwidth,keepaspectratio]{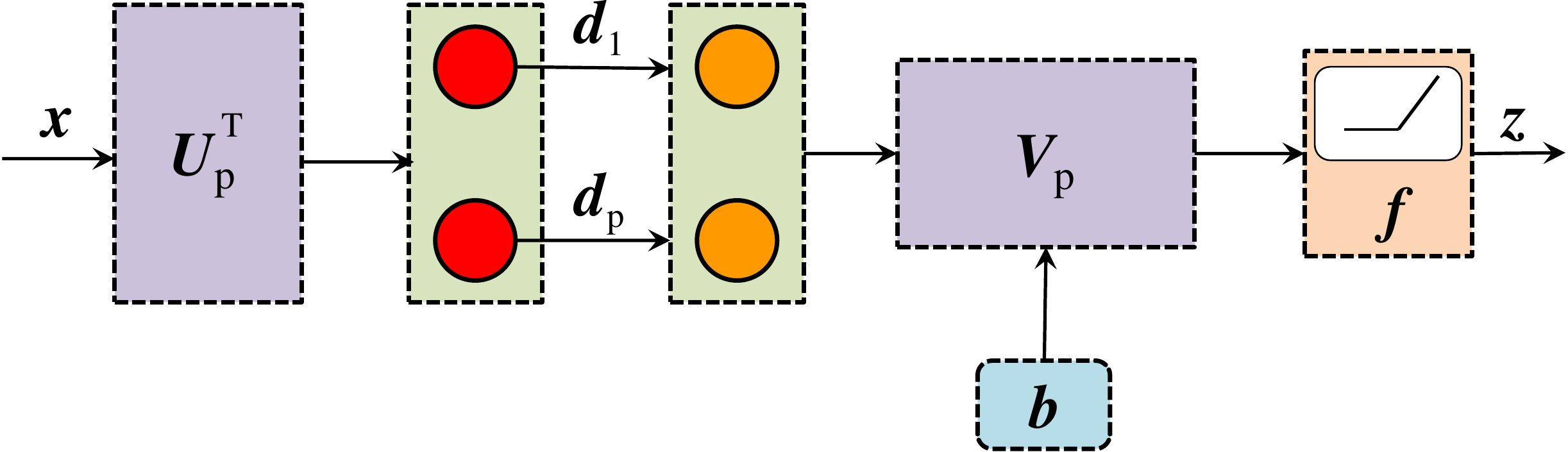}
\caption{\footnotesize \textbf{Low-rank factorization using Stiefel Layers.}  
The structure realizes the function $\Vec{z} = f\Big(\Mat{U}_p\Mat{D}_p\Mat{V}_p^T\Vec{x} + \Vec{b}\Big)$ with 
$\Mat{U}_p^T\Mat{U}_p = \Mat{V}_p^T\Mat{V}_p = \mathbf{I}_p$ and $\Mat{D}_p = \DIAG(d_{1},\cdots,d_{p})$.}
\label{fig:net_simplification}
\end{figure}

The discussion above suggests replacing an fc layer with the structure shown in Fig.~\ref{fig:net_simplification}, which in turn
reduces the number of parameters from $n_1 \times n_2$ to $(n_1  + n_2 - p + 2) \times p$. In practice,
after approximating the fc layer, one opts for fine-tuning the resulting network. 
We will show in our experiments that by adapting the above strategy, we can improve the classification accuracy of 
VGG-M~\cite{Simonyan_2014} while reducing its parameters substantially.

\begin{remark}[Unstructured Low-rank Approximation]
We note that a low-rank approximation to $\Mat{W}$ can be written in the form of $\Mat{W} = \Mat{G}\Mat{H}^T$ with
$\Mat{G} \in \mathbb{R}^{n_1 \times p}$ and $\Mat{H} \in \mathbb{R}^{n_2 \times p}$. As such, it may 
seem redundant to enforce orthogonality as discussed above.  
However, the decomposition in the form 
$\Mat{G}\Mat{H}^T$ is invariant to the right action of $\mathcal{O}(p)$, meaning that changing $\Mat{G}$ and $\Mat{H}$ to 
$\Mat{G}\Mat{R}$ and $\Mat{H}\Mat{R}$ with $\Mat{R} \in \mathcal{O}(p)$ leads to the same solution. 
Such invariances make the optimization difficult (especially Newton-like solvers) as shown by the recent work of 
Mishra \etal~\cite{Mishra_2014}.
The take home message from~\cite{Mishra_2014} is that the Euclidean geometry is not the right choice for addressing such decompositions 
(see for example Fig.7 in~\cite{Mishra_2014}), which makes our contribution  theoretically more appealing. 

\end{remark}

\section{Related Work}
\label{sec:related_work}

In this part, we review studies that target adding structures into deep networks.
Chan \etal propose PCANets, a hierarchical structure with orthogonal filter-banks to address the problem of 
image classification~\cite{Chan_TIP_2015}. The orthogonal filter-banks are learned by applying PCA over blocks 
extracted in the vicinity of the filters. 
The algorithm starts with constructing filters of the first layer directly from the training images. 
Once the first layer is constructed, the filters in the second layer are obtained by applying PCA over the responses
of the images gone through the first layer. This procedure is continued till all the layers are constructed (in practice 1-2 layers). 
Our development here enables us to close the learning loop, making end-to-end training with orthogonal filters 
possible which is the missing part in PCANets.

Costa and Fiori proposed to use principal component networks in a shallow structure for the task of image compression~\cite{Costa_2001}.
In particular, the principal component network is trained using the Generalized Hebbian Algorithm (GHA) of Sanger~\cite{Sanger_1989} 
which provides a way to obtain the top eigenvectors of a correlation matrix. Our work is simply a flexible extension of
GHA and principal component networks. Noting that GHA optimizes the PCA cost (minimizing the reconstruction error with the orthogonality constraint)
, it is easy to see that not only our work can  mimic the GHA but also any other form of dimensionality reduction in a deep structure. 
We refer the interested reader to the recent work 
of Cunningham and Ghahramani where various forms of dimensionality reduction are addressed by the geometry of the Stiefel 
manifold (though not in a deep structure)~\cite{Cunningham_JMLR_2015}.

Equipping deep nets with various structural forms has received 
growing attention lately~\cite{Yang_ICCV_2015_DFC,Ionescu_ICCV_2015}. 
Ionescu \etal extend the backpropagation algorithm to work with structured layers that perform matrix calculus~\cite{Ionescu_ICCV_2015}. 
One particular example is the normalized cut~\cite{Shi_2000} layers used for image segmentation~\cite{Ionescu_ICCV_2015}.
The structure within the input or the output space is also explored using convolutional neural networks~\cite{Chen2015,Fernando2016,Zheng2015}.
For instance, Chen~et~al.~\cite{Chen2015}  combine Markov Random Fields with deep learning to estimate complex representations while taking into account the dependencies between the output random
variables. 
Zheng~et~al.~\cite{Zheng2015} extended this idea and used Conditional Random Fields and convolutional neural networks to perform image segmentation.
Both these methods exploit the structure of output space and adapt the neural networks to accommodate such constraints.
Similarly, Fernando and Gould~\cite{Fernando2016} exploit the structure within the input data such as video sequences using convolutional neural networks.
However, in contrast, we enrich the neural networks by enforcing the structure within the network parameters and introduce novel version of back-propagation to accommodate constraints such as orthogonality\footnote{%
While preparing our paper, we came across the recent work of Huang and Van Gool~\cite{Huang_Arxiv_2016}. 
There, authors introduce SPDNets, a especial type of deep nets that accept Symmetric Positive Definite (SPD) matrices as inputs.
To keep the SPD structure across layers of the net, the authors make use of the Stiefel manifolds, though in a different context. 
}.


\section{Experimental Evaluation}
\label{sec:experiments}

In this section, we compare and contrast the proposed Stiefel layer trained by gBP on four different tasks, namely unsupervised feature learning, image classification, fine-grained image classification and network simplification.

\subsection{Unsupervised Learning}

As our first experiment, we tackle the problem of unsupervised feature learning using CMU-PIE 
face dataset~\cite{PIE_DB} (see Fig.~\ref{fig:PIE_samples} for examples). 
For this experiment, we resized images to $64 \times 64$ and used the gray-values as features. 
Images form nine different poses were used for training. Tests were performed on two unseen poses 
(poses that are not in the training set). This results in having 11253 images for training and 2811 for testing.

We trained three different structures of Denoising AutoEncoder \textbf{DAE}~\cite{Vincent_ICML_2008}, 
namely the conventional DAE,
the orthogonal DAE (\textbf{ODAE}) where we enforced the weights of the encoder to be orthogonal and
(\textbf{O$^2$DAE}) where both the encoder and the decoder enjoy orthogonal weights. To demonstrate the 
importance of orthogonality, we opt for the simplest network structure, making a a network with two fc layers, 
the encoder with size $4096 \times p$ and the decoder with size $p \times 4096$. We evaluated the performance of
DAE, ODAE and O$^2$DAE along PCA algorithm for various values of $p$ in Table~\ref{tab:pie_results}. 
To compute the classification accuracies, the output of the encoders after training was fed to a linear SVM.  
In all the experiments,   the learning rates were reduced from $10^{-4}$ to $10^{-6}$,
the batch sizes was set to 100, weight decay and the number of epochs were chosen as 0.0001 and 100, respectively.

Studying Table~\ref{tab:pie_results} reveals, both ODAE and O$^2$DAE outperform DAE for higher dimensionalities.
For $p = 64$, DAE is marginally better than ODAE but noticeably behind O$^2$DAE. For this case, PCA achieves the highest 
accuracy of 88.0\%. When $p$ is increased, PCA behaves erratically. Especially for $p = 1024$, PCA achieves the lowest accuracy 
among all the methods. DAE shows a consistent and saturating behavior when the dimensionality is increased. 
The performance of ODAE and O$^2$DAE enhances as $p$ is increased with a maximum of 94.6\% achieved by O$^2$DAE
for $p = 1024$.  This performance is significantly better than that of PCA and DAE. 

We note that while PCA reconstructs 
its input according to $\hat{\Vec{x}} = \Mat{W}_{pca}\Mat{W}_{pca}^T \Vec{x}$ with orthogonality constraint on $\Mat{W}_{pca}$, 
O$^2$DAE reconstructs its input according to $\hat{\Vec{x}} = \Mat{W}_{enc}\Mat{W}_{dec}^T \Vec{x}$ 
with orthogonality on $\Mat{W}_{enc}$ and $\Mat{W}_{dec}$. This is obviously a more flexible formulation, leading to 
higher accuracies as suggested in Table~\ref{tab:pie_results}.

\def \PIE_SIZE {0.08}
\begin{figure}[!tb]
      \centering
      \begin{subfigure}[b]{\PIE_SIZE \textwidth}
        	\includegraphics[width=\textwidth]{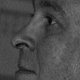}
      \end{subfigure}%
      \begin{subfigure}[b]{\PIE_SIZE \textwidth}
        	\includegraphics[width=\textwidth]{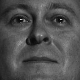}
      \end{subfigure}%
      \begin{subfigure}[b]{\PIE_SIZE \textwidth}
      		\includegraphics[width=\textwidth]{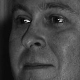}
      \end{subfigure}%
      \begin{subfigure}[b]{\PIE_SIZE \textwidth}
      		\includegraphics[width=\textwidth]{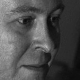}
      \end{subfigure}%
      \begin{subfigure}[b]{\PIE_SIZE \textwidth}
      		\includegraphics[width=\textwidth]{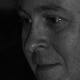}
      \end{subfigure}%
      \begin{subfigure}[b]{\PIE_SIZE \textwidth}
      		\includegraphics[width=\textwidth]{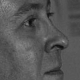}
      \end{subfigure}\\
      \begin{subfigure}[b]{\PIE_SIZE \textwidth}
        	\includegraphics[width=\textwidth]{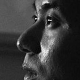}
      \end{subfigure}%
      \begin{subfigure}[b]{\PIE_SIZE \textwidth}
        	\includegraphics[width=\textwidth]{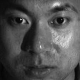}
      \end{subfigure}%
      \begin{subfigure}[b]{\PIE_SIZE \textwidth}
      		\includegraphics[width=\textwidth]{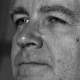}
      \end{subfigure}%
      \begin{subfigure}[b]{\PIE_SIZE \textwidth}
      		\includegraphics[width=\textwidth]{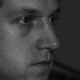}
      \end{subfigure}%
      \begin{subfigure}[b]{\PIE_SIZE \textwidth}
      		\includegraphics[width=\textwidth]{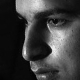}
      \end{subfigure}%
      \begin{subfigure}[b]{\PIE_SIZE \textwidth}
      		\includegraphics[width=\textwidth]{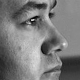}
      \end{subfigure}%
      \caption{\footnotesize Sample images from the CMU-PIE dataset~\cite{PIE_DB}.}
      \label{fig:PIE_samples}
\end{figure}

\begin{table}[h]
\centering
\begin{tabular}{|c|c|c|c|c|c|}
\hline
{\bf Method} 	&{\bf ~p = 64}	&{\bf ~p = 128} 	&{\bf ~p = 256}		&{\bf ~p = 512} 
&{\bf ~p = 1024}\\
\hline 
PCA    			
&\textbf{88.0}\% 		&83.2\% 	&90.9\% 	&86.1\% 		&78.9\%\\
DAE		        
				&82.4\% 	&85.3\% 	&85.9\%	 	&87.1\%				&87.4\%\\
ODAE    		
&82.1\%		&\textbf{87.7}\%		&\textbf{92.3}\%		&93.3\%				&94.1\%\\
O$^2$DAE		
&83.1\%		&86.8\%		&91.3\%		&\textbf{93.7}\%				&\textbf{94.6}\%\\
\hline 
\end{tabular}
\caption{\footnotesize Classification accuracies for \textbf{PCA}, Denoising AutoEncoder \textbf{(DAE)}, 
Orthogonal DAE \textbf{(ODAE)} (orthogonality is imposed on the encoder) and  
\textbf{O$^2$DAE} (orthogonality is imposed on both the encoder and the decodere) on the PIE dataset~\cite{PIE_DB}.}
\label{tab:pie_results}
\end{table}

\subsection{Supervised Learning}

As the second experiment, we consider the task of image classification using the classical LeNet~\cite{Lecun_1995} architecture.
The network is shipped with the MatConvNet package~\cite{matconvnet} and constitutes of 4 convolutional layers, 3 pooling layers
along one fc layer in the form \emph{input 
$\to$ conv($5\times5\times32$) $\to$ max-pooling($3\times3$)
$\to$ conv($5\times5\times32$) $\to$ ave-pooling($3\times3$)
$\to$ conv($5\times5\times64$) $\to$ ave-pooling($3\times3$)
$\to$ conv($4\times4\times64$)  
$\to$ fc($64\times10$) $\to$ 
output
}. 

The  orthogonal LeNet (o-LeNet) has the same structure as that of LeNet with one 
difference. The fc layer in the o-LeNet enjoys the orthogonality constraint on its weights.
We compare the performance of o-LeNet to that of LeNet   
using CIFAR10~\cite{CIFAR_DB},  CIFAR100~\cite{CIFAR_DB} and STL~\cite{STL_DB} datasets.

The CIFAR10 and CIFAR100 datasets~\cite{CIFAR_DB} are composed of $32 \times 32$
RGB images belonging to 10 and 100 different classes, respectively. Both datasets contain 50K training
images and 10K test images. 
The STL dataset~\cite{STL_DB} has 8k images for training and 5k for testing.
Images are RGB and belong to 10 different classes.
For the STL dataset, we downsampled the original images from $96 \times 96$ to $32 \times 32$.

To train the networks, we normalized all the datasets for contrast normalization and applied ZCA whitening.
In all the experiments,   the learning rates were reduced from $10^{-1}$ to $10^{-3}$,
the batch sizes was set to 100, weight decay and the number of epochs were chosen as 0.0005 and 250, respectively. 
As for data augmentation, we only considered image flipping during training.

Classification accuracies reported in Table~\ref{tab:lenet_results} show that the orthogonality constrain helps in all three cases.
While the improvements on the CIFAR10 and CIFAR100 datasets are substantial, it is  a totally different story for the STL dataset.
In this case, the o-LeNet outperforms LeNet by near 10 percentage points, suggesting that constraining the search space by orthogonality 
when fewer training samples are available is more important.

\begin{table}[h]
\centering
\begin{tabular}{|r|c|c|c|}
\hline
{\bf } 	&{\bf ~CIFAR10} &{\bf ~CIFAR100}	&{\bf ~STL}\\
\hline 
LeNet    			&79.6\%		&51.7\% 	&51.4\%\\
o-LeNet		        &\textbf{82.4}\%		&\textbf{54.7}\% 	&\textbf{62.1}\%\\
\hline 
\end{tabular}
\caption{\footnotesize Classification accuracies for \textbf{LeNet}~\cite{Lecun_1995}, 
and orthogonal LenNet \textbf{o-LeNet} on CIFAR10, CIFAR100~\cite{CIFAR_DB} and STL~\cite{STL_DB} datasets.}
\label{tab:lenet_results}
\end{table}

\subsection{Fine-Grained Image Classification}

We now incorporate the Stiefel layers into very deep networks such as VGG-VD~\cite{Simonyan_2014} and tackle the challenging problem of fine-grained image classification using three datasets, namely Birds~\cite{CUB200_DB}, aircrafts~\cite{AIRCRAFT_DB}, and cars~\cite{CARS196_DB}. 

The Birds dataset officially known as CUB-200-2011~\cite{CUB200_DB} dataset contains 11,788 images of 200 north American bird species. 
The FGVC-aircraft dataset~\cite{AIRCRAFT_DB} consists of 10,000 images of 100 aircraft variants.
The cars dataset~\cite{CARS196_DB} contains 16,185 images of 196 classes.
Both aircraft and cars datasets were introduced as a part of the FGComp 2013 challenge.
For all dataset, we resized images to $256 \times 256$ and followed the standard splits and evaluation protocols.
We evaluate our methods with Bounding Box (BB) information and without it (w/o~BB) using  
AlexNet~\cite{Krizhevsky_NIPS_2012}, VGG-F~\cite{Simonyan_2014}, VGG-M~\cite{Simonyan_2014} and VGG-VD~\cite{Simonyan_2014} networks trained on 
the ImageNet dataset~\cite{Imagenet}.

To assess the impact of the Stiefel layer, two different setups were considered.
In the first setup, we introduced the Stiefel layer just after the first fully connected layer.
We call this setup \emph{FC6-Stiefel}.
In the second setup, we placed the Stiefel layer after the second fully connected layer (\emph{FC7-Stiefel}).
In each case, we add the softmax classifier just after the Stiefel layer.

To fine-tune the networks, we initialized the classifier parameters using a linear SVM.
That is, we first extracted the activations of the respective network, followed by training an SVM using LibLinear package~\cite{LIBLINEAR}.
We then used the learned SVM to initialize the weights of the classifier.
For the generic networks and their Stiefel counterparts, we used the same learning rates (linearly decaying from $10^{-3}$ to $10^{-4}$) for fine tuning. The batch sizes,  weight decay and the number of epochs were set to 64, 100 and 0.0005, respectively.

We first use the Cars dataset (as it is the largest of all) to contrast our method against the standard fine-tuning 
using VGG-M and VGG-VD networks. For the Stiefel layer, we set the dimensionality of the layer to 200.
Results are reported in Table~\ref{tbl:cars.full}.
As it can be seen from Table~\ref{tbl:cars.full}, the Stiefel layer improves over the baselines by a significant margin in most of the cases.
In some instances the improvement is 12.8\% (w/o~BB, FC6 setup). In the follow up experiments, we only use Stiefel-FC6 network 
as it leads to better accuracies according to Table~\ref{tbl:cars.full}.

\begin{table}[h]
\centering
\begin{tabular}{|l|c|c|c|c|}
\hline
\multirow{2}{*}{Method} & \multicolumn{2}{c|}{\bf VGG-M} & \multicolumn{2}{c|}{\bf VGG-VD} \\ \cline{2-5}
   	& w/o~BB 	& BB  &  w/o~BB 	& BB 	\\ \hline
FC7-Baseline    & 61.2      	& 77.5 &  76.8	& 86.0	\\ 
FC7-Stiefel 	& \textbf{67.3}	& \textbf{79.4} & \textbf{81.8}	& \textbf{86.9}	\\ \hline 
FC6-Baseline    & 58.4		& 77.4 &  71.3	& 84.3	\\
FC6-Stiefel 	& \textbf{71.2}		& \textbf{82.2} &  \textbf{82.4}	& \textbf{87.9}	\\ \hline
\end{tabular}
\caption{\footnotesize VGG-M and VGG-VD results for Cars dataset.}
\label{tbl:cars.full}
\end{table}

As our next experiment, we evaluate the performance of the Stiefel-FC6 for fine-grained image classification over all the three datasets
using AlexNet~\cite{Krizhevsky_NIPS_2012}, VGG-M~\cite{Simonyan_2014} and VGG-VD~\cite{Simonyan_2014} networks.
We used a fixed dimensionality, 200 to be specific, for the Stiefel layer in all the experiments.
Table~\ref{tbl:finegrain.all} provides the classification accuracies for all three datasets, with and without bounding box information.

\begin{table}[t]
\centering
\begin{tabular}{|l|c|c|c|c|c|c|} \hline
\multirow{2}{*}{\bf Model} 
&  \multicolumn{2}{c|}{\bf Birds} & \multicolumn{2}{c|}{\bf Aircrafts} & \multicolumn{2}{c|}{\bf Cars} \\ \cline{2-7} 
					& w/o~BB& BB & w/o~BB& BB  	 &  w/o~BB	& BB 	\\ \hline
					
AlexNet-Baseline   			&   60.0\%   &   68.4\%		&   73.7\% 		& 	78.7\%   &   64.1\%   	&   79.0\% \\
AlexNet-Stiefel   			&   \bf 62.5\%   &   \bf 70.5\%		&   \bf 74.6\%   	&   \bf 80.3\%   &   \bf64.2\%   	&   \bf79.1\% \\
\hline
Vgg-F-Baseline   			&   61.8\%   &   69.0\%		&   73.2\% 		& 	79.9\%   &  65.0\%  		&   79.1\% \\
Vgg-F-Stiefel   			&   \bf64.4\%   &   \bf71.0\%		&   \bf74.3\%   	&   \bf81.3\%   &   \bf65.3\%  	&   \bf79.5\% \\
\hline
Vgg-M-Baseline   			&   65.6\%   &   70.8\%		&   76.3\% 		& 	81.4\%   &   61.2\%  	&   77.5\% \\
Vgg-M-Stiefel   			&   \bf67.9\%   &   \bf73.5\%		&   \bf78.0\%   	&  \bf 83.1\%   &   \bf71.2\%   	&   \bf82.2\% \\
\hline
Vgg-VD-Baseline   			&   75.3\%   &   79.3\%		&	83.6\% 		& 	86.9\%   &   76.8\%   	&   86.0\% \\  
Vgg-VD-Stiefel   			&   \bf76.0\%   &   \bf81.2\%		&	\bf84.3\%   	&   \bf87.1\%   &   \bf82.4\%   	&   \bf87.9\% \\  \hline
\end{tabular}
\caption{\footnotesize Fine-grained classification accuracies using our Stiefel layer.}
\label{tbl:finegrain.all}
\end{table}

Table~\ref{tbl:finegrain.all} reveals that our method obtains consistent improvement over the standard networks. 
Furthermore, at times, our results are  comparable with tailored deep models for fine-grained image classification
such as B-CNN~\cite{Lin2015}.
For example the B-CNN~\cite{Lin2015} obtains the best accuracy of 84.1\% for the Aircraft dataset while 
we obtain an accuracy of 84.3\% without benefiting from any fine-grained designs in the network architecture.
Furthermore, the recent work of~\cite{Krause2015}, a tailored method for fine-grained recognition, obtains 
an accuracy of 82.8\% on the Birds dataset. Our method, reaching the accuracy of 81.2\%, is marginally worse compared to~\cite{Krause2015}
while again not benefiting from  fine-grained techniques.

\begin{figure}[tb]
\centering
\includegraphics[width=0.9\linewidth]{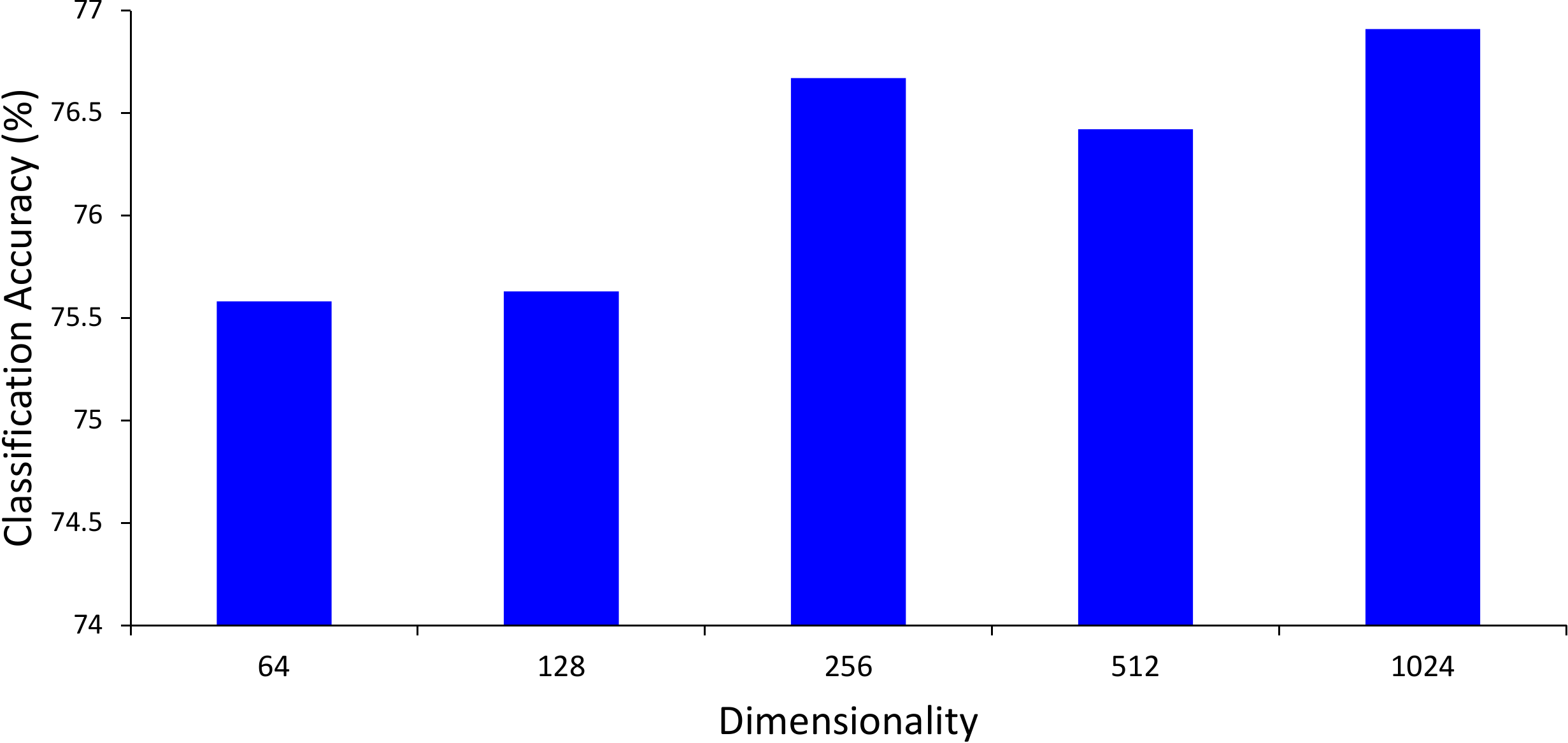}
\caption{\footnotesize Classification performance of the Stiefel layer with varying dimensionality on Birds dataset using VGG-VD network architecture.}
\label{fig:birds.dim.vgg.vd}    
\end{figure}

To further study the behavior of the Stiefel layer, we assess the impact of the dimensionality of the Stiefel layer on the classification accuracy. For this experiment and using the FC6-Stiefel model, we changed the dimensionality of the Stiefel layer 
in the range $[64, 128, 256, 512, 1024]$.
We report results using VGG-VD network and birds dataset without using bounding boxes.
Results are shown in Fig.~\ref{fig:birds.dim.vgg.vd}.
Fig.~\ref{fig:birds.dim.vgg.vd} shows that the classification accuracy increases up to 256 dimensions and seems to stabilize. 
Most interestingly, even for smaller dimensionality of 64, the Stiefel layer leads to a very reasonable classification accuracy of 75.6\%.

\subsection{Network Simplification}

As our final experiment, we evaluate the performance of the VGG-M network by simplifying its fc7 layer according to the 
development proposed in \textsection~\ref{sec:lr_layer_simplification}. For this experiment, we used the Cars dataset~\cite{CARS196_DB}.
Per Table~\ref{tbl:finegrain.all}, the classification accuracy of VGG-M network after fine-tuning on cars  
is 77.5\% (using bounding boxes). 

We systematically obtained a low-rank approximation  by preserving $\rho\%$ of the energy of the weight matrix of fc7 using SVD. 
Then we replaced the fc7 layer with its low-rank approximation, benefiting from two Stiefel 
layers (see \textsection~\ref{sec:lr_layer_simplification}).
We fine-tuned the resulting network for 50 epochs with learning rate reducing from 0.001 to 0.0001.
Table~\ref{tab:lr_layer_simplification} shows the classification accuracies for various values of the parameter $\rho$. 
We also report the number of parameters
of the resulting low-rank combination (two Stiefel layers) in Table~\ref{tab:lr_layer_simplification}. We note that the 
number of parameters of the fc7 layer is 16.7M. Interestingly, while the network is heavily downsized, its performance is improved.

\begin{table}[h]
\centering
\begin{tabular}{|r|c|c|c|c|c|}
\hline
{\bf } 	&{$\rho = 20$} &{$\rho = 30$}	&{$\rho = 40$}	&{$\rho = 50$}	&{$\rho = 60$}\\
\hline 
Accuracy    			&79.9\%		&81.4\% 	&82.0\%		&81.9\%		&81.6\%\\ \hline
No. Params.		        &269k		&472k 		&745k		&1175k		&1994k\\
\hline 
\end{tabular}
\caption{\footnotesize Low-rank approximation of the fc7 layer in VGG-M using Cars dataset~\cite{CARS196_DB}.}
\label{tab:lr_layer_simplification}
\end{table}


\section{Conclusions and Future Work}

In this work we introduced a generalized form of the BackPropagation (BP) algorithm~\cite{Bryson_1969,Rumelhart_1986} 
that can preserve structural forms on the weights of a network. This is achieved by bringing ideas from the Riemannian geometry 
and optimization on matrix manifolds to generalize the BP method. We also introduced the Stiefel layer, a layer with orthogonal weights. We empirically showed
that Stiefel layers boost the classification accuracies in several image classification tasks using different network architectures.

In the future, we plan to make use of our generalized BP method to add other forms of structures to a network. 
In particular, we are keen to explore the positive definite and subspace constraints in designing deep networks. 

\balance
\bibliography{references}

\end{document}